\pdfoutput=1

\documentclass[11pt]{article}

\usepackage[preprint]{acl}

\usepackage{times}
\usepackage{latexsym}

\usepackage[T1]{fontenc}

\usepackage[utf8]{inputenc}

\usepackage{microtype}

\usepackage{inconsolata}

\usepackage{graphicx}
\usepackage{amsmath}
\usepackage{booktabs}
\usepackage{tabularx}
\usepackage{float}
\usepackage{multirow}
\usepackage{xcolor}
\usepackage{url}

\usepackage{amsfonts,amssymb,amsthm}
\graphicspath{{figures/}}

\title{AutoTerm-SST: Adaptive Terminology Memory with Zero Session-Time Setup for Streaming Speech Translation}

\author{
  Jiaxuan Luo\textsuperscript{1} \quad
  Siqi Ouyang\textsuperscript{2} \quad
  Lei Li\textsuperscript{2}\thanks{\ \ Corresponding author.} \\
  \textsuperscript{1}Johns Hopkins University \\
  \textsuperscript{2}Carnegie Mellon University \\
  \texttt{jluo50@jhu.edu}, \texttt{\{siqiouya,leili\}@andrew.cmu.edu}
}

\begin{document}
\maketitle

\begin{abstract}
Retrieval-augmented speech translation improves rare terminology, yet systems
such as Retrieval-Augmented Simultaneous Speech Translation, or RASST, expect
a user glossary before each session. Merging all domains into one glossary
removes this step, but fixed retrieval and prompt budgets let irrelevant and conflicting
entries crowd out useful terms as the catalog grows. We present AutoTerm-SST,
an interactive streaming speech translator without glossary upload or domain
selection. It partitions an offline terminology catalog into topic slices.
A small language model scores each slice against recent partial translations.
A sticky topic filter forms a posterior from the scores, and retrieval
searches only the most probable slices under an unchanged budget. On a
4.7-hour English-to-Chinese stream alternating ACL talks with oncology
lectures, AutoTerm-SST reaches 87.7\% terminology accuracy, on par with a
reference told each talk's domain at 87.5\%, and it beats a
one-million-entry merged glossary by 3.3 accuracy points and 2.31 BLEU. On ten-domain
RealSI streams, the language-model router keeps the correct topic slice searchable in
89.7\% of audio chunks, versus 78.8\% for an embedding-similarity router with tuned
confirmation rules. The live interface shows active slices and retrieved term pairs with
each partial translation.
\end{abstract}

\section{Introduction}

Simultaneous speech translation, or SST, generates target-language text while
a speaker is still talking, serving technical talks, lectures, and live media.
Speech-capable large language models, or LLMs, now translate long-form audio directly
\citep{xu2025qwen3omni,ouyang2025infinisst}. Rare method and dataset names, clinical
concepts, and named entities remain hard to recover from acoustics alone. Glossary retrieval helps by matching speech to bilingual term pairs
and prompting the LLM with top matches before decoding.

\begin{figure*}[t]
\centering
\includegraphics[width=0.86\textwidth]{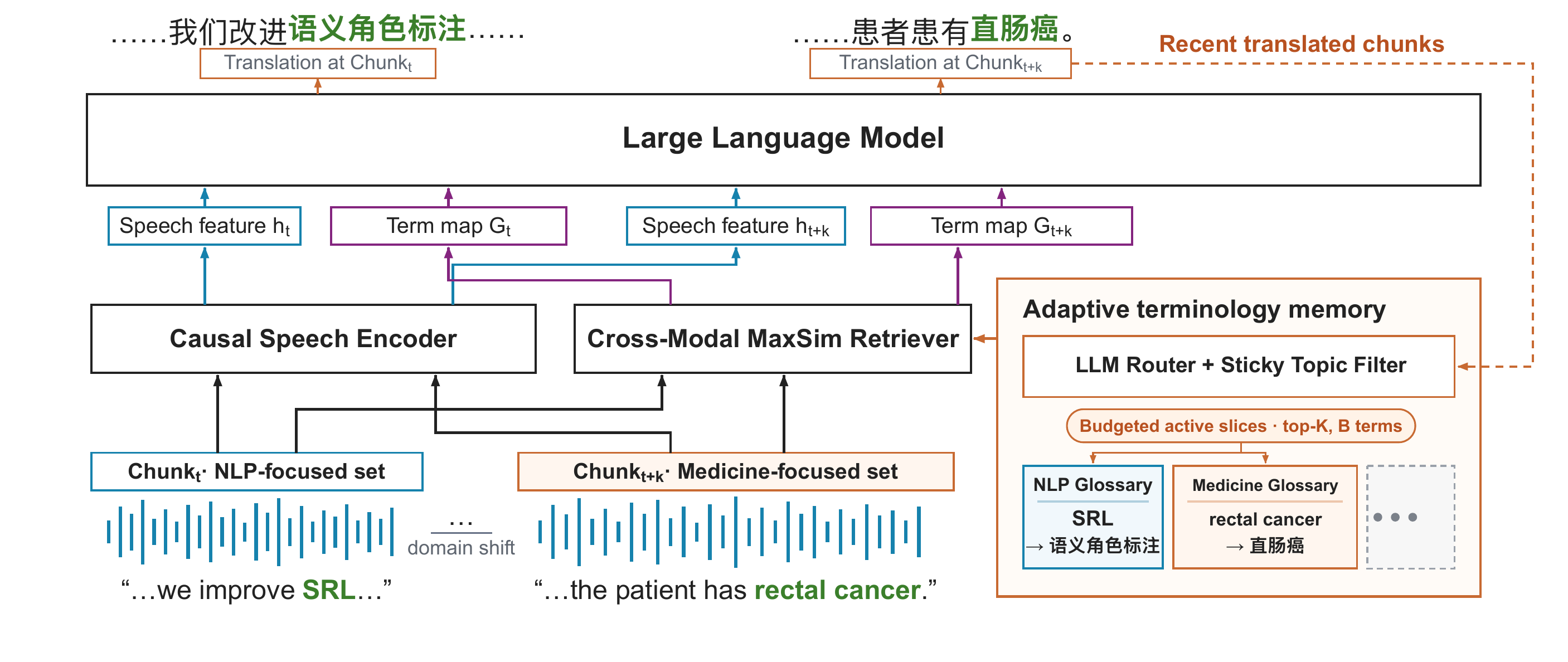}
\caption{AutoTerm-SST during a live topic transition from NLP to medicine.
The adaptive terminology memory, in orange, scores topic slices against recent translations on the
dashed arrow. It activates up to four slices with at most 4,000 terms. The MaxSim
retriever builds each chunk's term map $G_t$ from this working set alone. The LLM
translates speech features $h_t$ with $G_t$. The ellipsis marks
additional offline topic slices.}
\label{fig:architecture}
\end{figure*}

Existing retrieval-augmented SST systems need manual session setup or lose
accuracy as terminology catalogs grow. Retrieval-Augmented Simultaneous
Speech Translation, or RASST \citep{luo2026rasst}, improves terminology on
out-of-domain talks, but fixes the glossary before translation. Listeners
rarely know a live talk's exact domain in advance. Finding, uploading, or
selecting a glossary before every session adds friction. Merging all domains
into one glossary avoids manual selection and works at small scale. With
fixed counts of retrieval candidates and prompt term pairs, irrelevant and
conflicting mappings crowd out relevant terms as the catalog grows. In our
evaluation, increasing the size of a merged glossary from 100,000 to one million
entries brings terminology accuracy down from 87.8\% to 84.4\%, as
Table~\ref{tab:quality} shows.

The translation stream itself supplies the evidence to choose a glossary. The
RealSI streams~\citep{cheng2024clasi} change domains every 3.4 to 5.2 minutes. Our
4-billion-parameter language-model router reads the last four translated chunks and
ranks a new talk's glossary first after a median of 15.6 seconds. A streaming translator can therefore infer the domain from its own
output without asking the listener to declare it.

We present AutoTerm-SST, a system demo that moves glossary selection from user
setup into the streaming loop. Operators register offline topic slices, each
holding roughly 1,000 term pairs and a prototype, a short description of the
topic. During a session, a small LLM scores every prototype against recent
partial translations, a sticky hidden Markov filter accumulates these scores
into a posterior over topics, and the most probable slices, at most four, form
a working set inside a fixed term budget. Retrieval and decoding stay
unchanged, so the per-chunk budget stays constant as the catalog grows. In
summary, our contributions are threefold:
\begin{itemize}
\item Terminology-aware SST built on RASST, with a live interface that starts
without session setup and shows active slices and retrieved term pairs beside
every partial translation.
\item Training-free language-model routing with a sticky topic filter whose
only free parameter is the switch rate. The correct slice stays searchable
in 89.7\% of chunks on ten-domain RealSI streams. Embedding-similarity
routing with tuned confirmation rules reaches 78.8\%.
\item Evidence from a 4.7-hour stream alternating natural language processing,
or NLP, and oncology talks: AutoTerm-SST is on par with a known-domain reference and beats a
one-million-entry merged glossary by 3.3 points of terminology accuracy and
2.31 BLEU.
\end{itemize}

\section{Related work}

\paragraph{Streaming speech translation.}
Streaming SST balances quality and latency when decoding partial speech.
Prefix-to-prefix policies spell out this trade-off \citep{ma2019stacl},
attention-based policies refine read and write decisions
\citep{papi2023alignatt,papi2024streamatt}, and SimulEval standardizes
evaluation \citep{ma2020simuleval}. Speech-language models broaden direct SST
architectures \citep{seamless2023,zhang2024streamspeech,xu2025qwen3omni}.
AutoTerm-SST adds terminology management and an inspection layer to an
existing streaming model.

\paragraph{Terminology-aware translation.}
Terminology-aware methods assume the relevant glossary is known before decoding. Speech
translation struggles with rare terms and named entities \citep{gaido2021moby}.
Prior work uses user dictionaries via constrained decoding or training-time
exposure \citep{hokamp2017lexically,dinu2019training}. Retrieval-augmented
generation offers a non-parametric alternative
\citep{lewis2020retrieval,khandelwal2021nearest} that RASST applies to
streaming LLM decoding \citep{luo2026rasst}. AutoTerm-SST selects topic slices
online, removing pre-session glossary setup. Like interactive translation
tools \citep{dandekar2024translation}, it shows its choices and uses a
continuous-batching engine \citep{kwon2023efficient,zheng2024sglang}.

\section{System overview}
\label{sec:system}

Figure~\ref{fig:architecture} shows AutoTerm-SST's online terminology-selection
loop around RASST, whose models stay unchanged. RASST translates 1.92-second
audio chunks with a speech encoder, a cross-modal term retriever, and the
Qwen3-Omni LLM. The encoder maps chunk $t$ to features $h_t$, the retriever
builds a term map $G_t$, and the LLM translates $h_t$ with $G_t$ in its prompt.
Without a source transcript from RASST, the retriever matches speech directly
to glossary source terms by maximum similarity, or MaxSim, between overlapping
speech windows and pre-encoded term vectors. Terms scoring above 0.78 become
candidates. The retriever adds up to ten bilingual term pairs to $G_t$.
AutoTerm-SST changes only retrieval scope. Its router uses each translation
to select up to four topic slices for later searches.

Keeping adaptation outside the translation model makes the system pluggable.
One streaming protocol supports a retrieval-free translator, fixed-glossary
RASST, and AutoTerm-SST. Each partial translation sends the retrieved pairs,
their source slices, and the working set to the interface.

\section{Adaptive terminology memory}
\label{sec:adaptive-memory}

\begin{figure*}[t]
  \centering
  \begin{minipage}{.97\textwidth}
    \centering
    \includegraphics[height=.376\linewidth]{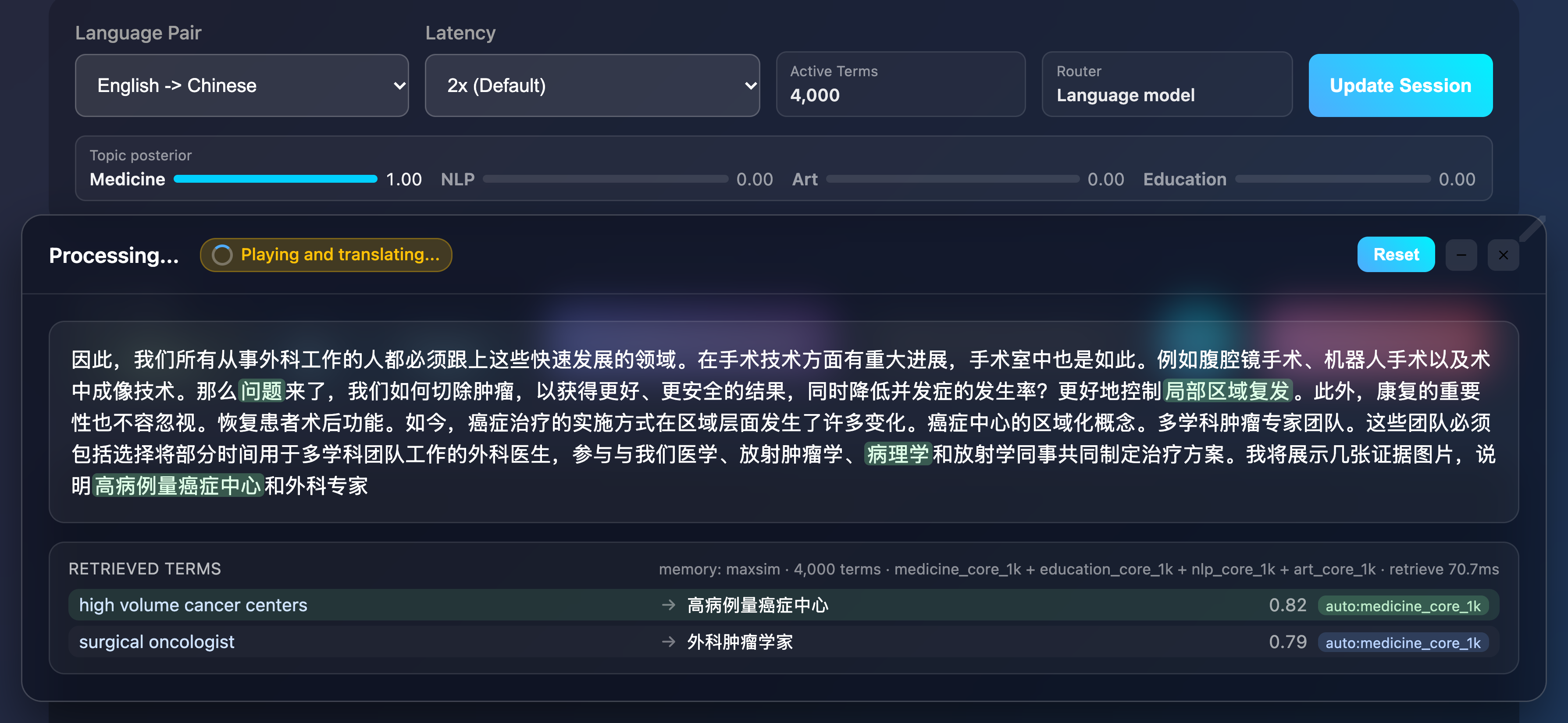}\hfill
    \includegraphics[height=.376\linewidth]{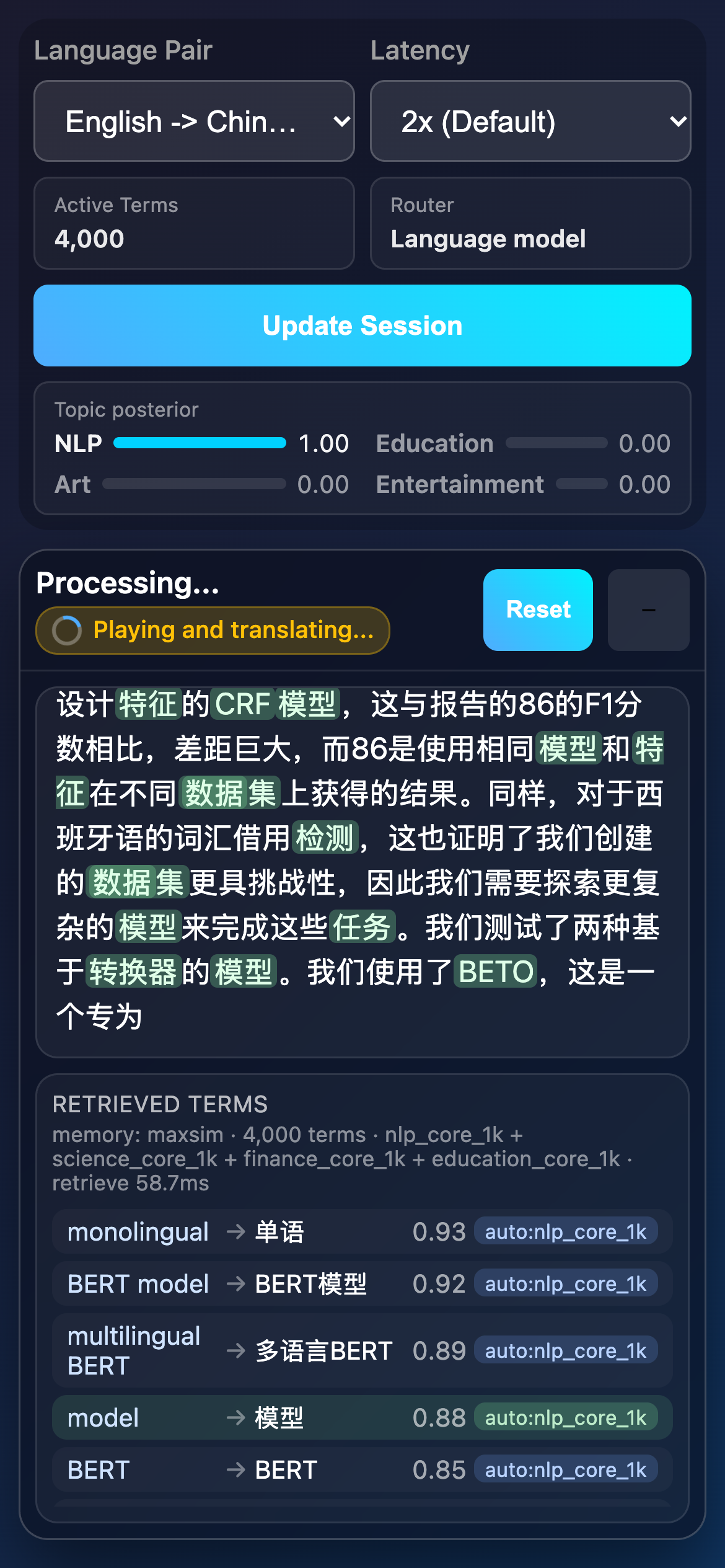}
    \caption{The English-to-Chinese interface replaying one session of the ACL-ESO
    stream. Left: an oncology lecture 19 minutes in, with the medicine slice leading
    the posterior. Right: the installable phone app during the first ACL talk,
    6 minutes in, with the NLP slice leading. Probabilities come from a replay of the
    recorded translations, and highlights mark retrieved terms.}
    \label{fig:ui}
  \end{minipage}
\end{figure*}

Separating an offline catalog from a small working set lets AutoTerm-SST
cover broad domains while keeping per-chunk search bounded. Each topic slice holds
about 1,000 terms with translations, a prebuilt MaxSim index of source terms, and a
prototype. The prototype gives a short English topic summary with example terms and
its Chinese translation. Different topic slices can assign different target
translations to the same source term.

\paragraph{Offline catalog.}
Adding a domain touches only the catalog manifest. An operator registers a
slice with one entry holding its label, term count, prototype, and index path. The router reads only the manifest and
prototypes, and the retriever skips every index outside the working set.

\paragraph{Language-model routing.}
At each point, the router asks a small instruction-tuned LLM which slice
matches the talk. The speech LLM translates audio directly without a source
transcript, so routing reads target-language output, which becomes more
topic-specific with context. At chunk $t$, the router concatenates the last
$w=4$ translated chunks into context $c_t$ and lists each slice's prototype
under a letter in a multiple-choice prompt. The prototype $P_d$ of slice $d$
gives a one-line description, then an English sentence naming its core topics
and representative terms. The 4-billion-parameter
Qwen3-4B-Instruct-2507~\citep{yang2025qwen3} serves as router LLM $\pi$. It
scores slice $d$ by the probability of its letter as the first answer token.
Catalog order $\sigma$ and its reverse $\bar\sigma$ put each slice at
mirrored positions. The mean offsets early or late position preferences and gives
\begin{equation}
  o_t(d)=\tfrac12\bigl[\pi(\ell_\sigma(d)\mid c_t)
  +\pi(\ell_{\bar\sigma}(d)\mid c_t)\bigr],
\end{equation}
where $\ell_\sigma(d)$ is the letter of $d$ under order $\sigma$. Each update
takes two forward passes and works without generating tokens or training the model.

\paragraph{Topic filtering.}
A sticky hidden Markov model combines noisy scores into a stable posterior $b_t$ over
the $N$ catalog slices. Between updates, the topic stays in its slice with
probability $1-\rho$ or moves uniformly to one of the other $N-1$ slices. With $o_t$
as the observation likelihood, the posterior update is
\begin{equation}
\begin{aligned}
  b_t(d)\propto{}&\Bigl[(1-\rho)\,b_{t-1}(d)\\
  &+\tfrac{\rho}{N-1}\bigl(1-b_{t-1}(d)\bigr)\Bigr]\,o_t(d)^{\beta_t}.
\end{aligned}
\end{equation}
To account for overlapping contexts, the exponent $\beta_t=\min(1,n_t/w)$ weights
each update by its share of new context. Here $n_t$ counts the chunks translated
since the previous update. The switch rate $\rho$ is
the filter's only free parameter. We set $\rho=10^{-3}$. In the replays of Section~\ref{sec:evaluation}, top-four coverage varies by at
most 0.5 points across the switch rates tested. Routing is causal: a decision at chunk $t$ affects only later chunks.

\paragraph{Working-set update.}
The working set holds the most probable slices so outgoing and incoming topics
can coexist during transitions. Every five seconds, the router scores every
catalog slice, updates the posterior, and fills the set by posterior rank up
to $K=4$ slices and $B=4{,}000$ searchable terms. Retrieval skips lower-ranked
slices, and stale slices leave as their rank falls. The complete router
configuration appears in Appendix~\ref{app:config}.

\paragraph{Fixed evidence budget.}
Active slices share one retrieval and prompt budget. The retriever splits
100 candidate slots equally among them, removes exact duplicates, reranks the
rest globally, and prompts with up to ten pairs that clear 0.78. AutoTerm-SST
thus gains by fitting more relevant mappings within the same fixed budget.

\paragraph{Non-blocking updates.}
A background worker keeps routing off the decoding path. It scores changed session
contexts in one batch. Each chunk retrieves from the working set at decode start.
Finished updates apply from the next chunk. Likely slice indexes are preloaded in the
background. In replays, applying every update one chunk late lowers coverage by
at most 0.33 points.

\section{Demo interface}
\label{sec:demo-interface}

AutoTerm-SST serves listeners of specialized live speech who cannot prepare a
glossary for every session, such as conference attendees, students in
technical courses, and clinicians following medical lectures. Users start
as they would with an ordinary streaming translator. They choose a language
pair and start a microphone or file stream without uploading a glossary or
selecting a domain.

\paragraph{Walkthrough.}
Sessions follow topic changes without manual control. Partial translations
appear immediately. As context builds, the top status row lists active topic
slices and their posterior probabilities. The evidence panel shows the pairs
retrieved for each output. After a topic change, outgoing and incoming topic slices
briefly share the working set. Figure~\ref{fig:ui} shows one session at two
points. During the ACL talk, the phone app shows the NLP slice leading the
posterior, and its evidence panel lists retrieved NLP pairs such as BERT model.
During the oncology lecture, the medicine slice leads, and the translation
renders high volume cancer centers with its retrieved target term.

\paragraph{Visible terminology evidence.}
Highlighting and an evidence panel make terminology behavior transparent. As
Figure~\ref{fig:ui} shows, the interface highlights retrieved terms in the
translation and lists the corresponding bilingual term pairs. The status row
shows the active topic slices, their posterior probabilities, and the active term count. The
evidence-panel header lists the active slices and retrieval latency. Viewers can tell
whether retrieval missed a mistranslated term, drew it from an irrelevant slice, or
supplied a term the decoder ignored.

\paragraph{Extensibility.}
Adding a topic requires one glossary file without a restart. Registration
validates pairs, builds the index, and writes the manifest entry. Prototypes
that GPT-5.5 writes from each glossary, with evaluation terms removed, keep
top-four coverage within 0.8 points of the hand-written ones, as
Appendix~\ref{app:router} reports.
Sessions retain the catalog snapshot from the start of each session.

\paragraph{Availability.}
The \href{https://github.com/luojiaxuan/autoterm-sst}{GitHub repository}
hosts the MIT-licensed implementation, a one-command launcher, and a screencast.
The hosted entry page is \url{https://luojiaxuan.github.io/autoterm-sst/}.
The web interface installs as a phone app on iOS and Android.
The live demo runs the RASST retriever alongside Qwen3-Omni, which it
serves with continuous batching in vLLM.

\section{Evaluation}
\label{sec:evaluation}

We test whether AutoTerm-SST matches a reference told each talk's domain
without user setup, avoids the losses of large merged glossaries with a fixed
working-set budget, and tracks frequent topic changes across a ten-domain stream.

\begin{figure*}[t]
  \centering
  \includegraphics[width=0.9\textwidth]{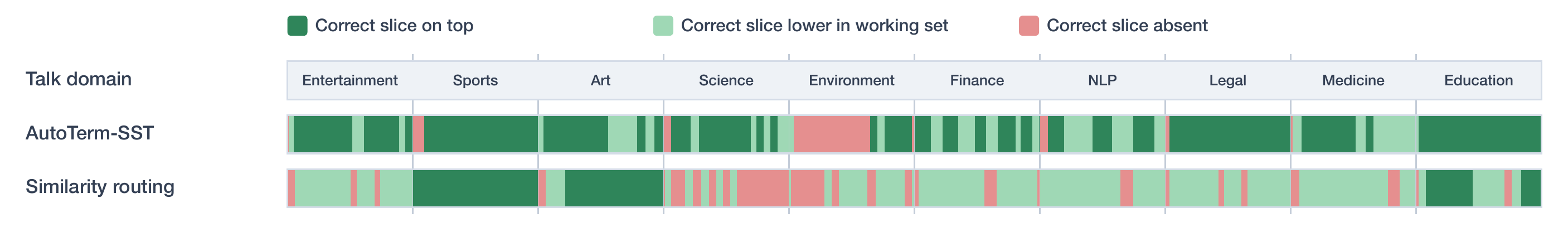}
  \caption{Routing in the first of three RealSI streams. The ten equal-width talks
  each have one domain. Per chunk, each row shows whether the correct topic slice
  ranks first, ranks lower in the working set, or is absent. AutoTerm-SST's
  language-model router ranks the correct topic slice first within 21 seconds in nine
  of the ten talks. The similarity router puts the correct topic slice in first
  place in three of the talks.}
  \label{fig:routing}
\end{figure*}

\subsection{Experimental setting}

\paragraph{Data.}
For English-to-Chinese translation quality, a continuous 4.7-hour ACL-ESO
stream alternates five English NLP talks from ACL 60/60 dev, with human
references and annotated terminology~\citep{salesky-etal-2023-evaluating}, and five
long-form oncology lectures from the European School of Oncology, or ESO,
test set~\citep{iranzosanchez2025eso}. For ESO, we use RASST's model-generated Chinese
references and manually verified terminology~\citep{luo2026rasst}.
The NLP and medicine slices contain the benchmark glossaries, so routing
determines whether gold terms are searchable. For ten-domain routing, we
arrange the ten English-to-Chinese RealSI talks~\citep{cheng2024clasi} in three
random orders. The 45-minute streams contain 30 talk instances in total.
Each talk lasts 3.4 to 5.2 minutes. Its human domain label marks one
of the ten slices as correct throughout.

\begin{table*}[t]
\centering
\footnotesize
\setlength{\tabcolsep}{4pt}
\begin{tabular}{lllrrrrrrr}
\toprule
& & & \multicolumn{3}{c}{TERM\_ACC} & & & & \\
\cmidrule(lr){4-6}
Setting & Input & Active & NLP & Medicine & All & BLEU & MT-BLEU & Prompt prec. & Pairs \\
\midrule
Known-domain-1k & Domain & 1k & 88.3 & 85.9 & 87.5 & 45.15 & 42.41 & 48.7 & 0.69 \\
AutoTerm-SST & None & 4k & 88.4 & 86.3 & 87.7 & 45.10 & 42.27 & 35.7 & 0.94 \\
Similarity routing & None & 4k & 88.4 & 86.0 & 87.6 & 45.39 & 42.62 & 35.7 & 0.94 \\
Merged-2k & Domains & 2k & 88.8 & 86.0 & 87.9 & 45.31 & 42.48 & 39.1 & 0.86 \\
Merged-10k & None & 9.9k & 88.6 & 87.0 & 88.1 & 45.11 & 42.33 & 27.1 & 1.26 \\
Merged-100k & None & 100k & 89.4 & 84.5 & 87.8 & 43.72$^\dagger$ & 40.93$^\dagger$ & 6.8 & 4.96 \\
Merged-1M & None & 1M & 87.1 & 78.9 & 84.4$^\dagger$ & 42.79$^\dagger$ & 39.95$^\dagger$ & 3.4 & 8.70 \\
\midrule
InfiniSST & None & 0 & 76.6 & 46.3 & 61.4 & 41.84 & 38.32 & 0.0 & 0.00 \\
\bottomrule
\end{tabular}
\caption{Results on the ACL-ESO stream. Input lists what the user supplies before a
session. Active gives the maximum entries searched per chunk. TERM\_ACC and prompt
precision are percentages. MT-BLEU masks glossary terms, and Pairs counts injected
pairs per chunk. $\dagger$: $p<0.05$ against AutoTerm-SST. InfiniSST runs without a
glossary. Its All, BLEU, and MT-BLEU scores average the two corpora with equal
weight.}
\label{tab:quality}
\end{table*}

\paragraph{Baselines.}
All configurations share RASST's model, chunks, retriever, threshold,
candidate slots, and prompt cap. They differ only in searchable entries.
The size in each baseline's name gives its searchable entries per chunk.
InfiniSST~\citep{ouyang2025infinisst} uses the same streaming model without
retrieval. Known-domain-1k searches only the 1,000-entry slice matching the
current talk and requires every talk's domain label. AutoTerm-SST routes over
ten 1,000-entry slices for NLP, medicine, and eight topics from Wikipedia and
Wikidata, with up to four active. The similarity router uses the same slices
and budgets and ranks by BGE-M3 embedding similarity. As
Appendix~\ref{app:config} details, it switches only when a candidate meets
fixed margins for three consecutive updates outside a 30-second cooldown.
Merged-2k requires knowing both domains and pools the NLP and medicine slices.
Merged-10k pools all ten slices, yielding 9,880 entries after removing exact
duplicates. Merged-1M adds other topics' entries to the NLP and medicine
glossaries. Its first 100,000 entries form Merged-100k. Merged glossaries
retain conflicting translations of a source term.

\paragraph{Metrics.}
RASST's terminology accuracy, or TERM\_ACC, counts the share of time-aligned
term occurrences with the reference translation in the aligned hypothesis.
The ACL-ESO stream has 2,047 evaluated occurrences, with 1,368 in NLP and 679 in
medicine. Prompt precision counts the share of injected pairs matching a gold
term in their source audio. RealSI coverage counts the share of chunks with the
correct slice in the working set. Translation quality uses corpus SacreBLEU and masked-term BLEU, detailed with paired permutation tests in Appendix~\ref{app:stats}. Intervals are 95\% talk-bootstrap bounds, written $a\pm h$ or $a^{+u}_{-l}$.

\subsection{Results}

\paragraph{Known-domain reference.}
Without user input, AutoTerm-SST performs on par with a known-domain reference.
AutoTerm-SST minus Known-domain-1k is $+0.2^{+0.8}_{-1.0}$ TERM\_ACC points
at $p=0.82$ and $-0.05^{+0.44}_{-0.55}$ BLEU at $p=0.85$.
Table~\ref{tab:quality} shows higher scores for the glossary runs than for
InfiniSST, which runs without retrieval.

\paragraph{Effect of catalog scale.}
Merged glossaries match routed retrieval up to 10k entries and lose quality beyond. Merged-2k and Merged-10k
lead AutoTerm-SST by at most 0.4 TERM\_ACC points at $p\ge0.34$ and match BLEU.
Merged-100k matches AutoTerm-SST in TERM\_ACC and trails by
$1.38^{+0.53}_{-0.70}$ BLEU at $p=0.010$. Merged-1M trails by
$3.3^{+1.5}_{-1.9}$ TERM\_ACC points at $p=0.031$ and $2.31^{+0.63}_{-0.82}$ BLEU at
$p=0.002$. Its TERM\_ACC loss concentrates in medicine, a 7.4-point gap.
Merged-1M injects pairs at nine times AutoTerm-SST's rate per chunk and at one
tenth of its prompt precision. On the 935 sentences with glossary-free references,
AutoTerm-SST's BLEU gain over Merged-1M is 1.85 points at $p=0.047$. All three gains hold when any single talk is left out.

\begin{table*}[t]
\centering
\small
\begin{tabular}{lrrrrrrrrr}
\toprule
& & \multicolumn{5}{c}{Topic switch every} & \multicolumn{2}{c}{Target language} & \\
\cmidrule(lr){3-7}\cmidrule(lr){8-9}
Router & ACL-ESO stream & 15 s & 30 s & 60 s & 120 s & 300 s & German & Japanese & Close domains \\
\midrule
Similarity & 97.0 & 82.1 & 90.5 & 89.3 & 89.6 & 93.5 & 97.2 & 97.8 & 52.2 \\
Language model & 99.0 & 100.0 & 98.2 & 97.9 & 97.4 & 99.3 & 97.9 & 99.0 & 82.7 \\
\bottomrule
\end{tabular}
\caption{Open-loop replays of saved translations by the two deployed routers. Updates
take effect on time at the live rate. Cells give the percentage of chunks with the
correct topic slice in the four-slice working set. The close-domain column gives the
percentage of gold term occurrences with their topic slice in the working set.}
\label{tab:router-diag}
\end{table*}

\begin{table}[t]
\centering
\small
\setlength{\tabcolsep}{4pt}
\begin{tabular}{lrrrr}
\toprule
Router & Coverage & Top-1 & To top & Never top \\
\midrule
AutoTerm-SST & 89.7 & 67.0 & 15.6 s & 1 \\
Similarity routing & 78.8 & 23.9 & 68.1 s & 19 \\
\bottomrule
\end{tabular}
\caption{Routing on 30 RealSI talk instances. Chunk shares track the correct
slice in the working set and in first place. The last columns give median time
to first place when reached and count talks that miss it.}
\label{tab:realsi}
\end{table}

\paragraph{Ten-domain routing.}
The language-model router tracks topic changes faster than similarity routing
with confirmation. RealSI coverage is 89.7\% against
78.8\% in Table~\ref{tab:realsi}, and the gap holds in each stream order in
Table~\ref{tab:realsi-orders}. The language-model router ranks a new talk's
slice first in a median of 15.6 seconds. Similarity routing misses first place
in 19 of 30 talk instances. Figure~\ref{fig:routing} shows the first order.
On two-domain ACL-ESO, both routers match in accuracy and cover the correct
slice in about 98\% of chunks. In Table~\ref{tab:router-diag}, the language-model router covers 82.7\% of gold
term occurrences with 24 close-domain slices, against 52.2\%, keeps 100.0\% coverage
when topics switch every 15 seconds, against 82.1\%, and leads on German and Japanese output.
When NLP and medicine alternate every 15 to 300 seconds, median time to first
place is 7.7 seconds with the language-model router and 27 to 46 seconds with
the similarity router.

\paragraph{Effect of the topic filter.}
The filter turns noisy per-update scores into a stable ranking without thresholds.
We test it in open-loop replays on the ACL-ESO stream with per-chunk updates, using
the saved translations described in Appendix~\ref{app:router}. Raw letter
probabilities rank the correct topic slice in the top four for 90.5\% of chunks,
and the filtered posterior reaches 98.7\%. Coverage stays between 98.4\% and
98.9\% as $\rho$ ranges from $10^{-5}$ to $10^{-1}$. Raw similarity covers
85.0\% of chunks with per-chunk updates. At the live rate, confirmation and
pinning give the similarity router 97.0\% coverage.

\paragraph{Processing speed.}
AutoTerm-SST processes audio about 4.5 times as fast as real time and matches
the real-time factor of the Known-domain-1k reference in Table~\ref{tab:speed}.
On the same GPUs, RASST retrieval alone took 0.042 seconds per chunk.
AutoTerm-SST emits its first translation 2.3 seconds after the audio starts.

\begin{table}[t]
\centering
\small
\begin{tabular}{lrrr}
\toprule
Setting & Retrieval & Generation & RTF \\
\midrule
Known-domain-1k & 0.059 & 0.359 & 0.22 \\
Similarity routing & 0.064 & 0.343 & 0.21 \\
AutoTerm-SST & 0.094 & 0.333 & 0.22 \\
\bottomrule
\end{tabular}
\caption{Mean seconds per 1.92 s chunk. One ACL talk uses three A6000 GPUs,
two for translation and one for retrieval and routing. Retrieval counts
speech encoding and routing. RTF divides compute time by audio time.}
\label{tab:speed}
\end{table}

\section{Conclusion}

AutoTerm-SST routes from partial translations to offline topic slices for
terminology-aware streaming translation without glossary upload or domain
selection. It matches a known-domain reference, avoids losses from a
million-entry merged glossary, and tracks topic changes across ten domains.
Scaling the router from tens of slices to thousands, with an embedding
shortlist or a topic hierarchy ahead of the letter prompt, is a natural next step.

\section*{Ethics statement}
We evaluate on the published ACL 60/60, ESO test set, and RealSI corpora,
using RealSI under CC BY 4.0 without redistributing its audio.

\section*{Acknowledgments}
The authors used Claude, GPT, and Gemini as utilities for coding, data
processing, and editing. They reviewed all outputs and take full responsibility.

\bibliography{custom}

\begin{thebibliography}{25}
\providecommand{\natexlab}[1]{#1}

\bibitem[{Chen et~al.(2024)Chen, Xiao, Zhang, Luo, Lian, and
  Liu}]{chen-etal-2024-m3}
Jianlyu Chen, Shitao Xiao, Peitian Zhang, Kun Luo, Defu Lian, and Zheng Liu.
  2024.
\newblock \href {https://doi.org/10.18653/v1/2024.findings-acl.137}
  {{M3}-embedding: Multi-linguality, multi-functionality, multi-granularity
  text embeddings through self-knowledge distillation}.
\newblock In \emph{Findings of the Association for Computational Linguistics:
  ACL 2024}, pages 2318--2335, Bangkok, Thailand. Association for Computational
  Linguistics.

\bibitem[{Cheng et~al.(2024)Cheng, Huang, Ko, Li, Peng, Xu, and
  Zhang}]{cheng2024clasi}
Shanbo Cheng, Zhichao Huang, Tom Ko, Hang Li, Ningxin Peng, Lu~Xu, and Qini
  Zhang. 2024.
\newblock Towards achieving human parity on end-to-end simultaneous speech
  translation via {LLM} agent.
\newblock \emph{arXiv preprint arXiv:2407.21646}.

\bibitem[{Dandekar et~al.(2024)Dandekar, Xu, Xu, Ouyang, and
  Li}]{dandekar2024translation}
Chinmay Dandekar, Wenda Xu, Xi~Xu, Siqi Ouyang, and Lei Li. 2024.
\newblock \href {https://arxiv.org/abs/2410.10861} {Translation canvas: An
  explainable interface to pinpoint and analyze translation systems}.
\newblock \emph{arXiv preprint arXiv:2410.10861}.

\bibitem[{Dinu et~al.(2019)Dinu, Mathur, Federico, and
  Al-Onaizan}]{dinu2019training}
Georgiana Dinu, Prashant Mathur, Marcello Federico, and Yaser Al-Onaizan. 2019.
\newblock \href {https://doi.org/10.18653/v1/P19-1294} {Training neural machine
  translation to apply terminology constraints}.
\newblock In \emph{Proceedings of the 57th Annual Meeting of the Association
  for Computational Linguistics}, pages 3063--3068, Florence, Italy.
  Association for Computational Linguistics.

\bibitem[{Gaido et~al.(2021)Gaido, Rodr{\'i}guez, Negri, Bentivogli, and
  Turchi}]{gaido2021moby}
Marco Gaido, Susana Rodr{\'i}guez, Matteo Negri, Luisa Bentivogli, and Marco
  Turchi. 2021.
\newblock Is {``}moby dick{''} a whale or a bird? named entities and
  terminology in speech translation.
\newblock In \emph{Proceedings of the 2021 Conference on Empirical Methods in
  Natural Language Processing}, Online and Punta Cana, Dominican Republic.
  Association for Computational Linguistics.

\bibitem[{Hokamp and Liu(2017)}]{hokamp2017lexically}
Chris Hokamp and Qun Liu. 2017.
\newblock \href {https://doi.org/10.18653/v1/P17-1141} {Lexically constrained
  decoding for sequence generation using grid beam search}.
\newblock In \emph{Proceedings of the 55th Annual Meeting of the Association
  for Computational Linguistics (Volume 1: Long Papers)}, pages 1535--1546,
  Vancouver, Canada. Association for Computational Linguistics.

\bibitem[{Iranzo-S{\'a}nchez et~al.(2025)Iranzo-S{\'a}nchez,
  Santamar{\'i}a-Jord{\`a}, Mas-Moll{\`a}, Garc{\'e}s D{\'i}az-Mun{\~i}o,
  Iranzo-S{\'a}nchez, Jorge, Silvestre-Cerd{\`a}, Gim{\'e}nez, Civera, Sanchis,
  and Juan}]{iranzosanchez2025eso}
Jorge Iranzo-S{\'a}nchez, Jaume Santamar{\'i}a-Jord{\`a}, Gerard Mas-Moll{\`a},
  Gon{\c{c}}al~V. Garc{\'e}s D{\'i}az-Mun{\~i}o, Javier Iranzo-S{\'a}nchez,
  Javier Jorge, Joan~Albert Silvestre-Cerd{\`a}, Adri{\`a} Gim{\'e}nez, Jorge
  Civera, Albert Sanchis, and Alfons Juan. 2025.
\newblock \href {https://doi.org/10.1016/j.artmed.2025.103147} {Speech
  translation for multilingual medical education leveraged by large language
  models}.
\newblock \emph{Artificial Intelligence in Medicine}, 166:103147.

\bibitem[{Khandelwal et~al.(2021)Khandelwal, Fan, Jurafsky, Zettlemoyer, and
  Lewis}]{khandelwal2021nearest}
Urvashi Khandelwal, Angela Fan, Dan Jurafsky, Luke Zettlemoyer, and Mike Lewis.
  2021.
\newblock Nearest neighbor machine translation.
\newblock In \emph{International Conference on Learning Representations}.

\bibitem[{Kwon et~al.(2023)Kwon, Li, Zhuang, Sheng, Zheng, Yu, Gonzalez, Zhang,
  and Stoica}]{kwon2023efficient}
Woosuk Kwon, Zhuohan Li, Siyuan Zhuang, Ying Sheng, Lianmin Zheng, Cody~Hao Yu,
  Joseph~E. Gonzalez, Hao Zhang, and Ion Stoica. 2023.
\newblock \href {https://doi.org/10.1145/3600006.3613165} {Efficient memory
  management for large language model serving with {PagedAttention}}.
\newblock In \emph{Proceedings of the {ACM} {SIGOPS} 29th Symposium on
  Operating Systems Principles}, pages 611--626. Association for Computing
  Machinery.

\bibitem[{Lewis et~al.(2020)Lewis, Perez, Piktus, Petroni, Karpukhin, Goyal,
  K{\"u}ttler, Lewis, Yih, Rockt{\"a}schel, Riedel, and
  Kiela}]{lewis2020retrieval}
Patrick Lewis, Ethan Perez, Aleksandra Piktus, Fabio Petroni, Vladimir
  Karpukhin, Naman Goyal, Heinrich K{\"u}ttler, Mike Lewis, Wen-tau Yih, Tim
  Rockt{\"a}schel, Sebastian Riedel, and Douwe Kiela. 2020.
\newblock \href
  {https://proceedings.neurips.cc/paper/2020/hash/6b493230205f780e1bc26945df7481e5-Abstract.html}
  {Retrieval-augmented generation for knowledge-intensive {NLP} tasks}.
\newblock In \emph{Advances in Neural Information Processing Systems},
  volume~33, pages 9459--9474.

\bibitem[{Luo et~al.(2026)Luo, Ouyang, Xu, and Li}]{luo2026rasst}
Jiaxuan Luo, Siqi Ouyang, Jiaxing Xu, and Lei Li. 2026.
\newblock \href {https://arxiv.org/abs/2601.22777} {{RASST}:
  Retrieval-augmented simultaneous speech translation}.
\newblock \emph{arXiv preprint arXiv:2601.22777}.

\bibitem[{Ma et~al.(2019)Ma, Huang, Xiong, Zheng, Liu, Zheng, Zhang, He, Liu,
  Li, Wu, and Wang}]{ma2019stacl}
Mingbo Ma, Liang Huang, Hao Xiong, Renjie Zheng, Kaibo Liu, Baigong Zheng,
  Chuanqiang Zhang, Zhongjun He, Hairong Liu, Xing Li, Hua Wu, and Haifeng
  Wang. 2019.
\newblock \href {https://doi.org/10.18653/v1/P19-1289} {{STACL}: Simultaneous
  translation with implicit anticipation and controllable latency using
  prefix-to-prefix framework}.
\newblock In \emph{Proceedings of the 57th Annual Meeting of the Association
  for Computational Linguistics}, pages 3025--3036, Florence, Italy.
  Association for Computational Linguistics.

\bibitem[{Ma et~al.(2020)Ma, Dousti, Wang, and Pino}]{ma2020simuleval}
Xutai Ma, Mohammad~Javad Dousti, Changhan Wang, and Juan Pino. 2020.
\newblock \href {https://doi.org/10.18653/v1/2020.emnlp-demos.19} {{SimulEval}:
  An evaluation toolkit for simultaneous translation}.
\newblock In \emph{Proceedings of the 2020 Conference on Empirical Methods in
  Natural Language Processing: System Demonstrations}, pages 144--150, Online.
  Association for Computational Linguistics.

\bibitem[{Matusov et~al.(2005)Matusov, Leusch, Bender, and
  Ney}]{matusov-etal-2005-evaluating}
Evgeny Matusov, Gregor Leusch, Oliver Bender, and Hermann Ney. 2005.
\newblock \href {https://aclanthology.org/2005.iwslt-1.19/} {Evaluating machine
  translation output with automatic sentence segmentation}.
\newblock In \emph{Proceedings of the Second International Workshop on Spoken
  Language Translation}, Pittsburgh, Pennsylvania, USA.

\bibitem[{Ouyang et~al.(2025)Ouyang, Xu, and Li}]{ouyang2025infinisst}
Siqi Ouyang, Xi~Xu, and Lei Li. 2025.
\newblock \href {https://aclanthology.org/2025.findings-acl.157/} {{InfiniSST}:
  Simultaneous translation of unbounded speech with large language model}.
\newblock In \emph{Findings of the Association for Computational Linguistics:
  ACL 2025}. Association for Computational Linguistics.

\bibitem[{Papi et~al.(2024)Papi, Gaido, Negri, and
  Bentivogli}]{papi2024streamatt}
Sara Papi, Marco Gaido, Matteo Negri, and Luisa Bentivogli. 2024.
\newblock {StreamAtt}: Direct streaming speech-to-text translation with
  attention-based audio history selection.
\newblock In \emph{Proceedings of the 62nd Annual Meeting of the Association
  for Computational Linguistics (Volume 1: Long Papers)}. Association for
  Computational Linguistics.

\bibitem[{Papi et~al.(2023)Papi, Negri, and Turchi}]{papi2023alignatt}
Sara Papi, Matteo Negri, and Marco Turchi. 2023.
\newblock {AlignAtt}: Using attention-based audio-translation alignments as a
  guide for simultaneous speech translation.
\newblock In \emph{Proceedings of INTERSPEECH 2023}.

\bibitem[{Papineni et~al.(2002)Papineni, Roukos, Ward, and
  Zhu}]{papineni-etal-2002-bleu}
Kishore Papineni, Salim Roukos, Todd Ward, and Wei-Jing Zhu. 2002.
\newblock \href {https://doi.org/10.3115/1073083.1073135} {{BLEU}: a method for
  automatic evaluation of machine translation}.
\newblock In \emph{Proceedings of the 40th Annual Meeting of the Association
  for Computational Linguistics}, pages 311--318, Philadelphia, Pennsylvania,
  USA. Association for Computational Linguistics.

\bibitem[{Post(2018)}]{post-2018-call}
Matt Post. 2018.
\newblock \href {https://doi.org/10.18653/v1/W18-6319} {A call for clarity in
  reporting {BLEU} scores}.
\newblock In \emph{Proceedings of the Third Conference on Machine Translation:
  Research Papers}, pages 186--191, Brussels, Belgium. Association for
  Computational Linguistics.

\bibitem[{{Qwen Team}(2025)}]{xu2025qwen3omni}
{Qwen Team}. 2025.
\newblock \href {https://arxiv.org/abs/2509.17765} {{Qwen3-Omni} technical
  report}.
\newblock \emph{arXiv preprint arXiv:2509.17765}.

\bibitem[{Salesky et~al.(2023)Salesky, Darwish, Al-Badrashiny, Diab, and
  Niehues}]{salesky-etal-2023-evaluating}
Elizabeth Salesky, Kareem Darwish, Mohamed Al-Badrashiny, Mona Diab, and Jan
  Niehues. 2023.
\newblock \href {https://doi.org/10.18653/v1/2023.iwslt-1.2} {Evaluating
  multilingual speech translation under realistic conditions with
  resegmentation and terminology}.
\newblock In \emph{Proceedings of the 20th International Conference on Spoken
  Language Translation (IWSLT 2023)}, pages 62--78, Toronto, Canada (in-person
  and online). Association for Computational Linguistics.

\bibitem[{{Seamless Communication}(2023)}]{seamless2023}
{Seamless Communication}. 2023.
\newblock \href {https://arxiv.org/abs/2312.05187} {Seamless: Multilingual
  expressive and streaming speech translation}.
\newblock \emph{arXiv preprint arXiv:2312.05187}.

\bibitem[{Yang et~al.(2025)}]{yang2025qwen3}
An~Yang et~al. 2025.
\newblock \href {https://arxiv.org/abs/2505.09388} {{Qwen3} technical report}.
\newblock \emph{arXiv preprint arXiv:2505.09388}.

\bibitem[{Zhang et~al.(2024)Zhang, Fang, Guo, Ma, Zhang, and
  Feng}]{zhang2024streamspeech}
Shaolei Zhang, Qingkai Fang, Shoutao Guo, Zhengrui Ma, Min Zhang, and Yang
  Feng. 2024.
\newblock {StreamSpeech}: Simultaneous speech-to-speech translation with
  multi-task learning.
\newblock In \emph{Proceedings of the 62nd Annual Meeting of the Association
  for Computational Linguistics (Volume 1: Long Papers)}. Association for
  Computational Linguistics.

\bibitem[{Zheng et~al.(2024)Zheng, Yin, Xie, Sun, Huang, Yu, Cao, Kozyrakis,
  Stoica, Gonzalez, Barrett, and Sheng}]{zheng2024sglang}
Lianmin Zheng, Liangsheng Yin, Zhiqiang Xie, Chuyue Sun, Jeff Huang, Cody~Hao
  Yu, Shiyi Cao, Christos Kozyrakis, Ion Stoica, Joseph~E. Gonzalez, Clark
  Barrett, and Ying Sheng. 2024.
\newblock {SGLang}: Efficient execution of structured language model programs.
\newblock In \emph{Advances in Neural Information Processing Systems},
  volume~37.

\end{thebibliography}

\clearpage
\appendix

\begin{table*}[t]
\centering
\footnotesize
\setlength{\tabcolsep}{3pt}
\begin{tabular}{lrrrrrrrrrrrr}
\toprule
& \multicolumn{4}{c}{TERM\_ACC} & \multicolumn{4}{c}{BLEU} & \multicolumn{4}{c}{MT-BLEU} \\
\cmidrule(lr){2-5}\cmidrule(lr){6-9}\cmidrule(lr){10-13}
Baseline & Diff. & Low & High & $p$ & Diff. & Low & High & $p$ & Diff. & Low & High & $p$ \\
\midrule
\multicolumn{13}{l}{\emph{All sentences}} \\
Known-domain-1k & $+0.20$ & $-0.83$ & $+0.99$ & 0.82 & $-0.05$ & $-0.60$ & $+0.39$ & 0.85 & $-0.13$ & $-0.75$ & $+0.35$ & 0.61 \\
Similarity routing & $+0.10$ & $-0.92$ & $+0.93$ & 0.94 & $-0.29$ & $-0.75$ & $+0.01$ & 0.14 & $-0.35$ & $-0.98$ & $-0.02$ & 0.13 \\
Merged-2k & $-0.20$ & $-1.19$ & $+0.56$ & 0.83 & $-0.21$ & $-0.57$ & $+0.24$ & 0.36 & $-0.21$ & $-0.61$ & $+0.24$ & 0.46 \\
Merged-10k & $-0.39$ & $-1.05$ & $+0.18$ & 0.34 & $-0.01$ & $-0.25$ & $+0.21$ & 0.96 & $-0.06$ & $-0.30$ & $+0.18$ & 0.63 \\
Merged-100k & $-0.10$ & $-1.27$ & $+0.82$ & 0.97 & $+1.38$ & $+0.68$ & $+1.91$ & 0.010 & $+1.35$ & $+0.64$ & $+1.75$ & 0.006 \\
Merged-1M & $+3.27$ & $+1.38$ & $+4.79$ & 0.031 & $+2.31$ & $+1.49$ & $+2.94$ & 0.002 & $+2.32$ & $+0.87$ & $+2.96$ & 0.002 \\
\midrule
\multicolumn{13}{l}{\emph{The 935 sentences whose references contain no glossary term}} \\
Known-domain-1k &  & & &  & $-0.42$ & $-0.97$ & $+0.07$ & 0.17 &  & & &  \\
Merged-100k &  & & &  & $+1.35$ & $+0.64$ & $+1.83$ & 0.023 &  & & &  \\
Merged-1M &  & & &  & $+1.85$ & $+0.52$ & $+3.01$ & 0.047 &  & & &  \\
\bottomrule
\end{tabular}
\caption{Paired differences on the ACL-ESO stream. We subtract each baseline score
from AutoTerm-SST's score. Positive values favor AutoTerm-SST. Low and High are the
95\% stratified talk-bootstrap confidence bounds. The $p$ column gives the exact
two-sided talk-level permutation $p$-value.
TERM\_ACC differences are in percentage points.}
\label{tab:stats}
\end{table*}

\section{Statistical comparisons}
\label{app:stats}

\paragraph{Translation metrics.}
mWERSegmenter~\citep{matusov-etal-2005-evaluating} resegments the hypotheses
to the StreamLAAL references~\citep{papi2024streamatt} before we compute
corpus SacreBLEU~\citep{papineni-etal-2002-bleu,post-2018-call}.
Masked-term BLEU, or MT-BLEU, repeats scoring with every evaluated target
term masked in both hypothesis and reference.

\paragraph{Paired tests.}
Table~\ref{tab:stats} reports AutoTerm-SST minus each baseline. The exact
permutation test swaps outputs between systems within each talk and recomputes
the pooled metric for all 1,024 combinations. The smallest
attainable two-sided $p$-value is 0.002. Significance claims rest on this test.
With ten talks, the percentile bootstrap can disagree with it near the boundary.
For MT-BLEU against similarity routing, the interval is below $0$ while $p=0.13$.
Confidence intervals come from 4,000 talk-level bootstrap resamples
stratified by corpus, with five ACL and five ESO talks. We write them as
$a\pm h$ when symmetric and $a^{+u}_{-l}$ otherwise.
When any single talk is removed, AutoTerm-SST still leads Merged-1M by at least
2.57 TERM\_ACC, 2.04 BLEU, and 1.70 MT-BLEU points.

\section{Router configuration}
\label{app:config}

Beyond the shared settings in Table~\ref{tab:config}, the language-model router
adds the switch rate $\rho$ and the similarity router adds five settings. The
latter uses BGE-M3~\citep{chen-etal-2024-m3} to embed the last 128 tokens of $c_t$
and each prototype with its Chinese translation. It pins the confirmed slice
and switches after a candidate leads both the runner-up and the confirmed slice
by fixed margins for three consecutive updates outside a 30-second cooldown.
We tuned the thresholds on four short segments from the evaluation talks before
running the complete streams. Each router ran on its own server because
the updates use wall-clock time.

\begin{table}[!ht]
\centering
\small
\setlength{\tabcolsep}{4pt}
\begin{tabular}{lr}
\toprule
Setting & Value \\
\midrule
\multicolumn{2}{l}{\emph{Shared}} \\
Audio chunk & 1.92 s \\
Translation context $w$ & 4 chunks \\
Maximum active slices $K$ & 4 \\
Active term budget $B$ & 4,000 \\
Routing update interval & 5 s \\
Retrieval score threshold & 0.78 \\
Candidate slots & 100 \\
Prompt cap & 10 pairs \\
\midrule
\multicolumn{2}{l}{\emph{Language-model router}} \\
Router LLM & Qwen3-4B-Instruct-2507 \\
Switch rate $\rho$ & $10^{-3}$ \\
\midrule
\multicolumn{2}{l}{\emph{Similarity router}} \\
Warm-up before first switch & 6 s \\
Cooldown after a switch & 30 s \\
Lead over runner-up & 0.15 \\
Lead over confirmed slice & 0.10 \\
Consecutive agreeing updates & 3 \\
\bottomrule
\end{tabular}
\caption{Router and retrieval settings. Leads are differences between normalized
similarity scores. The similarity router switches only when a candidate
maintains both leads across the required consecutive updates.}
\label{tab:config}
\end{table}

\section{Router prompt and prototypes}

Figure~\ref{fig:router-prompt} shows the router prompt.
Table~\ref{tab:router-prototypes} lists the hand-written prototypes for the
ACL-ESO runs, and the RealSI streams use the generated prototypes described in
Appendix~\ref{app:router}.

\begin{figure}[t]
\setlength{\fboxsep}{6pt}
\fbox{\parbox{\dimexpr\linewidth-2\fboxsep-2\fboxrule\relax}{\microtypesetup{protrusion=false}\small\raggedright
A live speech translator is translating a talk from English into
\texttt{\{language\}}. Below is its most recent \texttt{\{language\}} output,
followed by terminology glossaries, one per topic. Which glossary best matches
the topic of the talk at this point?\par\medskip
Recent translation:\\\texttt{\{context\}}\par\medskip
Glossaries:\\\texttt{\{options\}}\par\medskip
Answer with the letter of the best glossary only.}}
\caption{Prompt for the language-model router.}
\label{fig:router-prompt}
\end{figure}

\begin{table}[!t]
\centering
\small
\begin{tabularx}{\columnwidth}{@{}l@{\quad}X@{}}
\toprule
Slice & English prototype \\
\midrule
art & Painting, sculpture, museums, architecture, and art-history terms. Art, painting, sculpture, museums, artists, portraits, tapestries, frescoes, galleries, and art history. \\
education & School, teaching, curriculum, assessment, and learning terms. Education, schools, students, teachers, classrooms, curriculum, assessment, colleges, and universities. \\
entertainment & Film, television, music, gaming, and media terms. Entertainment, films, television, music, video games, actors, directors, consoles, and popular media. \\
environment & Climate, ecology, conservation, pollution, and disaster terms. Environment, climate, ecology, pollution, conservation, floods, storms, natural disasters, forests, and rivers. \\
finance & Market, instrument, accounting, monetary, and trading terms. Finance, markets, banking, investment, accounting, business, brands, prices, customers, and the economy. \\
legal & Law, court, contract, regulation, and liability terms. Law, courts, criminal cases, lawyers, contracts, legislation, legal rights, judgments, and appeals. \\
medicine & Clinical, disease, drug, procedure, and biomedical terms. Healthcare, clinical medicine, patients, diagnosis, treatment, drugs, disease, and public health. \\
nlp & NLP, speech, translation, ML, dataset, and benchmark terms. Natural language processing, speech translation, machine learning, software systems, servers, databases, and load balancing. \\
science & Biology, chemistry, physics, experiments, and research terms. Science, biology, chemistry, physics, experiments, cells, organisms, species, molecules, and evolution. \\
sports & Teams, tournaments, athletes, matches, and competition terms. Sports, football, basketball, teams, players, coaches, matches, leagues, tournaments, and championships. \\
\bottomrule
\end{tabularx}
\caption{English prototypes for the ACL-ESO router.}
\label{tab:router-prototypes}
\end{table}

\section{Router diagnostics}
\label{app:router}

Open-loop replays give both deployed routers the same saved ACL-ESO
translations and use the context in Section~\ref{sec:adaptive-memory}.
As in live runs, scores refresh at most once every 5 seconds of audio.
Each refresh applies the similarity router's confirmation rule and pin,
or one language-model filter step. The similarity router reproduces 19 topic
switches from the live run when given the original emission times.

\paragraph{Short topic shifts.}
The posterior tracks topics alternating every 15 to 300 seconds. For each
interval in Table~\ref{tab:router-diag}, one 40-segment stream alternates saved NLP
and medicine translations. In each stream, at least half the segments start
with the incoming slice active for the language-model router.
Section~\ref{sec:evaluation} reports time to first place for both routers.

\paragraph{Close domains.}
Finer slices widen the routing gap. The same 4B model splits each of NLP
and medicine into eight subtopics. It proposes subtopics from 400 terms
outside the evaluation glossaries and assigns terms by letter probability.
The other eight topics stay in the catalog. Table~\ref{tab:router-diag}
tracks working-set coverage of the topic slices for 2,047 gold term
occurrences. Selecting the best four slices per talk in hindsight gives
97.3\% coverage of the gold term occurrences.

\paragraph{Other target languages.}
Routing coverage carries over to German and Japanese in replays of the same
talks with RASST translations and only the prompt language changed.

\paragraph{Generated prototypes and late updates.}
GPT-5.5 wrote one prototype per slice from its glossary with evaluation
terms removed. With per-chunk updates on ACL-ESO, the generated prototypes
lower the language-model router's top-four coverage to 97.9\%. RealSI uses
them because its NLP talk shares topics with the hand-written NLP prototype.
Table~\ref{tab:realsi-orders} gives both routers' coverage and top-1 accuracy
by stream order. Closed-loop runs in Section~\ref{sec:evaluation} scored
updates before the next chunk. The deployed worker can apply an update one
chunk late when audio arrives faster than real time. We replay the settings in
Table~\ref{tab:router-diag} with each update one chunk late.
Section~\ref{sec:adaptive-memory} reports the coverage change.

\begin{table}[H]
\centering
\small
\begin{tabular}{lrrrr}
\toprule
& \multicolumn{2}{c}{Language model} & \multicolumn{2}{c}{Similarity} \\
\cmidrule(lr){2-3}\cmidrule(lr){4-5}
Order & Coverage & Top-1 & Coverage & Top-1 \\
\midrule
1 & 91.2 & 69.4 & 78.6 & 21.3 \\
2 & 88.6 & 63.2 & 79.2 & 26.9 \\
3 & 89.3 & 68.4 & 78.5 & 23.6 \\
\bottomrule
\end{tabular}
\caption{Routing by RealSI stream order. Coverage and top-1 accuracy report
chunk percentages with the correct slice in the working set or ranked first,
respectively.}
\label{tab:realsi-orders}
\end{table}

\end{document}